\documentclass[11pt,a4paper]{article}
\usepackage[hyperref]{naaclhlt2018}
\usepackage{times}
\usepackage{latexsym}
\usepackage{url}
\usepackage{amsmath}
\usepackage{amssymb}
\usepackage{csquotes}
\usepackage[disable]{todonotes}   
\usepackage{graphicx}
\usepackage{pgfplots}
\usepackage{adjustbox}
\usepackage{soul}

\newcommand{\note}[4][]{\todo[author=#2,color=#3,size=\scriptsize,fancyline,caption={},#1]{#4}} 
\newcommand{\katharina}[2][]{\note[#1]{Katharina}{green!40}{#2}}
\newcommand{\manuel}[2][]{\note[#1]{Manuel}{orange!40}{#2}}
\newcommand{\ivan}[2][]{\note[#1]{Ivan}{blue!40}{#2}}

\newcommand{\Katha}[2][]{\katharina[inline,#1]{#2}\noindent}
\newcommand{\Manuel}[2][]{\manuel[inline,#1]{#2}\noindent}

\aclfinalcopy 
\def\aclpaperid{94} 

\newcommand{\ptheta}{p_{{\boldsymbol \theta}}}

\title{Fortification of Neural Morphological Segmentation Models for Polysynthetic Minimal-Resource Languages}

\newcommand*\samethanks[1][\value{footnote}]{\footnotemark[#1] }
\author{  
  Katharina Kann\thanks{*The first two authors contributed equally.}\\ 
 Center for Information and\\Language Processing\\
  LMU Munich, Germany\\
  \texttt{kann@cis.lmu.de} \\
  \And
  Manuel Mager\samethanks\\
  Instituto de Investigaciones en\\Matem\'{a}ticas Aplicadas y en Sistemas\\
  Universidad Nacional Aut\'{o}noma de M\'{e}xico \\
  \texttt{mmager@turing.iimas.unam.mx} \\
  \AND
  Ivan Meza-Ruiz \\
  Instituto de Investigaciones en\\Matem\'{a}ticas Aplicadas y en Sistemas\\
  Universidad Nacional Aut\'{o}noma de M\'{e}xico \\
  \texttt{ivanvladimir@turing.iimas.unam.mx} \\
  \And
  Hinrich Sch\"{u}tze \\
  Center for Information and\\Language Processing\\
  LMU Munich, Germany\\
  \texttt{inquiries@cislmu.org} \\
}
\date{}

\pgfplotsset{compat=1.14}

\begin{document}
\maketitle
\begin{abstract}
Morphological segmentation for polysynthetic languages is challenging, because a word may consist of many 
individual morphemes and
training data can be extremely scarce.
Since neural sequence-to-sequence (seq2seq) models define the state of the art for morphological segmentation in high-resource settings and for (mostly) European languages, we first show that they also obtain competitive performance for Mexican polysynthetic languages in minimal-resource settings. We then propose two novel multi-task training approaches---one with, one without need for external unlabeled resources---, and two corresponding data augmentation methods, improving over the neural baseline for all languages. Finally, we explore cross-lingual transfer as a third way to fortify our neural model and show that we can train one single multi-lingual model for related languages while maintaining comparable or even improved performance, thus reducing the amount of parameters by close to $75\%$.
We provide our morphological segmentation datasets for Mexicanero, Nahuatl, Wixarika and Yorem Nokki for future research.
\end{abstract}

\section{Introduction}
Due to the advent of computing technologies to indigenous communities all over the world, 
natural language processing (NLP) applications for languages with limited computer-readable textual data are getting increasingly important. 
This contrasts with current research, which focuses strongly on approaches which require large amounts of training data, e.g., deep neural networks.
Those are not trivially applicable to \emph{minimal-resource} settings with less than $1,000$ available training examples.
We aim at closing this gap for morphological surface segmentation, the task of splitting a word into the surface forms of its smallest meaning-bearing units, its \emph{morphemes}. 

Recovering morphemes provides information about unknown words and is thus especially important for polysynthetic languages with a high morpheme-to-word ratio and a consequently large overall number of words. 
To illustrate how segmentation helps understanding unknown multiple-morpheme words, consider an example in this paper's language of writing: 
even if the word \emph{unconditionally} did not appear in a given training corpus,
its meaning could still be derived from a combination of its morphs \emph{un}, \emph{condition}, \emph{al} and \emph{ly}.

Due to its importance for down-stream tasks \cite{creutz2007morph,dyer2008generalizing}, segmentation has been tackled in many different ways, 
considering unsupervised \cite{creutz-lagus:2002:ACL02-MPL}, supervised \cite{ruokolainen-EtAl:2013:CoNLL-2013} and semi-supervised settings \cite{ruokolainen2014}. 
Here, we add three new questions to this line of research:
(i) Are data-hungry neural network models applicable to segmentation of polysynthetic languages in minimal-resource settings?
(ii) How can the performance of neural networks for surface segmentation be improved if we have only unlabeled or no external data at hand?
(iii) Is cross-lingual transfer for this task possible between related languages?
The last two questions are crucial: While for many languages it is difficult to obtain the number of annotated examples used in earlier work on (semi-)supervised methods, a limited amount might still be obtainable.    

We experiment on four polysynthetic Mexican languages: Mexicanero, Nahuatl, Wixarika and Yorem Nokki (details in \S\ref{sec:langs}).
The datasets we use are, as far as we know, the first computer-readable datasets annotated for morphological segmentation in those languages.

Our experiments show that neural seq2seq models
perform on par with or better than other strong baselines
for our polysynthetic languages in a minimal-resource setting.
However, we further propose two novel multi-task approaches and two new data augmentation methods. 
Combining them with our neural model yields up to $5.05\%$ absolute accuracy or $3.40\%$ F1 improvements over our strongest baseline.

Finally, following earlier work on cross-lingual knowledge transfer for seq2seq tasks \cite{TACL1081,kann-cotterell-schutze:2017:Long}, 
we investigate training one single model for all languages, while sharing parameters. The resulting model performs comparably to or better than the individual models, but requires only roughly as many parameters as one single model.

\paragraph{Contributions.} To sum up, we make the following contributions:
(i) we confirm the applicability of neural seq2seq models to morphological segmentation of polysynthetic languages in minimal-resource settings; (ii) we propose two novel multi-task training approaches and two novel data augmentation methods for neural segmentation models;
(iii) we investigate the effectiveness of cross-lingual transfer between related languages; and (iv) we provide morphological segmentation datasets for Mexicanero, Nahuatl, Wixarika and Yorem Nokki.

\section{Polysynthetic Languages}
\label{sec:langs}
 \begin{table}[t]
    \setlength{\tabcolsep}{3.pt} 
  \centering
  \begin{tabular}{ l c | l c | l c | l c }
  \multicolumn{2}{c|}{Mexicanero} &  \multicolumn{2}{c|}{Nahuatl} &  \multicolumn{2}{c|}{Wixarika} & \multicolumn{2}{c}{Yorem N.} \\\hline
frq.&m.       &frq.&      m.&frq.&m. &frq.& m.\\\hline\hline
                                       	
136 &       ni&155 &       o&327 &p+ &102 &  k\\
128 &       ki& 99 &      ni&230 &ne & 88 &  m\\
114 &       ti& 84 &      ti&173 &p  & 87 & ne\\
105 &        u& 81 &       k&169 &ti & 83 & ka\\
 70 &        s& 61 &      tl&167 &ka & 79 & ta\\
 44 &       mo& 59 &      mo& 98 &u  & 54 & po\\
 42 &       ka& 55 &       s& 97 &ta & 50 & e'\\
 39 &        a& 52 &      ki& 95 &a  & 36 & ye\\
 31 &     nich& 48 &       i& 92 &pe & 36 & su\\
 31 &      \$i& 43 &     tla& 91 &e  & 36 & ri\\
 24 &       ta& 39 &     'ke& 80 &r  & 34 &  a\\
 24 &        l& 34 &    nech& 74 &wa & 31 & me\\
 22 &tahtanili& 31 &      no& 69 &me & 30 & wa\\
 21 &       no& 27&      ya& 68& ni  & 30 & re\\
 17 &       ya& 27 &     tli& 68 &ke & 27 & na\\
 17 &        t& 24 &       x& 66 &eu & 24 & wi\\
 17 &       ke& 23 &tlanilia& 58 &ye & 24 & ßa\\
 17 &      ita& 23 &       e& 57 &ri & 23 & te\\
 16 &     piya& 21 &    tika& 52 &tsi& 20 & si\\
 15 &       an& 21 &       n& 52 &te & 16 &'wi\\

  \end{tabular}
  \caption{The most frequent morphs (\emph{m.}) together with their frequencies (\emph{frq.}) in our datasets.\label{tab:data3}}
\end{table}

Polysynthetic languages are morphologically rich languages which are highly synthetic, i.e., single words can be composed of many individual morphemes.
In extreme cases, entire sentences consist of only one single token, whereupon ``every argument of a predicate must be expressed by morphology on the word that contains that assigner''~\cite{Baker_1on}.
This property makes surface segmentation of 
polysynthetic languages at the same time complex and particularly relevant for further linguistic analysis.

In this paper, we experiment on four polysynthetic languages of the Yuto-Aztecan family \cite{baker1997complex},
with the goal of improving the performance of neural seq2seq models.
The languages will be described in the rest of this section.  

\paragraph{Mexicanero}
 is a Western Peripheral Nahuatl variant, spoken in the Mexican state of Durango by approximately one thousand people. This dialect is isolated from the rest of the other branches and has a strong process of Spanish stem incorporation, while also having borrowed some suffixes from that language \citep{vanhove2012morphologies}. It is common to see Spanish words mixed with Nahuatl agglutinations. In the following example we can see an intrasentencial mixing of Spanish (\textit{in uppercases}) and Mexicanero:
\begin{center}
u$\vert$ni$\vert$ye \textsc{malo} -- \emph{I was sick}
\end{center}

\Katha{I took out the alphabets. If you think they are important, just put them back in.}
\Manuel{Ok! :)}

\paragraph{Nahuatl}
\Katha{Do we want to put accents on Spanish words, e.g., San Luis Potos\'{i}?}
\manuel{I don't know. What do you think? }
 is a large subgroup of the Yuto-Aztecan language family, and, including all of its variants, the most spoken native language in Mexico. 
Its almost two million native speakers live mainly in Puebla, Guerrero, Hidalgo, Veracruz, and San Luis Potosi, but also in Oaxaca, Durango, Modelos, Mexico City, Tlaxcala, Michoacan, Nayarit and the State of Mexico. Three dialectical groups are known: Central Nahuatl, Occidental Nahuatl and Oriental Nahuatl. The data collected for this work belongs to the Oriental branch spoken by 70 thousand people in Northern Puebla.  

Like all languages of the Yuto-Aztecan family, Nahuatl is agglutinative and one word can consist of a combination of 
many different morphemes. Usually, the verb functions as the stem and gets extended by morphemes specifying, e.g., subject, patient, object or indirect object. The most common syntax sequence for Nahuatl is SOV. An example word is:
\begin{center}
o$\vert$ne$\vert$mo$\vert$kokowa$\vert$ya -- \emph{I was sick} 
\end{center}


\paragraph{Wixarika} 
is a language spoken in the states of Jalisco, Nayarit, Durango and Zacatecas in Central West Mexico by approximately fifty thousand people. It belongs to the Coracholan group of languages within the Yuto-Aztecan family. 
Wixarika has five vowels \{a,e,i,+\footnote{While linguists often use a dashed i (\st{i}) to denote this vowel, in practice almost all native speakers use a plus symbol (+). In this work, we choose to use the latter.},u\} with long and short variants.
An example for a word in the language is:
\begin{center}
ne$\vert$p+$\vert$ti$\vert$kuye$\vert$kai -- \emph{I was sick}
\end{center}
~\\
Like Nahuatl, it has an SOV syntax, with heavy agglutination on the verb. Wixarika is morphologically more complex than other languages from the same family, because it incorporates more information into the verb \citep{leza2006gramatica}. This leads to a higher number of morphemes per word as can also be seen in Table \ref{tab:data2}.  

\paragraph{Yorem Nokki} 
 is part of Taracachita subgroup of the Yuto-Aztecan language family.
 \Katha{Which dialect do we use?}
 \Manuel{Southern}
Its Southern dialect is spoken by close to forty thousand people in the Mexican states of Sinaloa and Sonora, 
while its Northern dialect has about twenty thousand speakers.
In this work, we consider the Southern dialect.
The nominal morphology of Yorem Nokki is rather simple, but,
like in the other Yuto-Aztecan languages, the verb is highly complex.
 \begin{table}[t]
    \setlength{\tabcolsep}{2.5pt} 
  \centering
  \begin{tabular}{l || c | c | c | c }
          &  Mexicanero & Nahuatl & Wixarika &Yorem N.\\\hline
    train &  427        & 540     & 665      &511     \\
    dev   &  106        & 134     & 176      &127     \\
    test  &  355        & 449     & 553      &425     \\ \hline
    total &  888        & 1123    & 1394     &1063    \\
  \end{tabular}
  \caption{Number of examples in the final data splits for all languages.\label{tab:data}}
\end{table} 
Its alphabet consists of 28 characters and contains 8 different vowels. An example verb is:
\begin{center}
ko'kore$\vert$ye$\vert$ne -- \emph{I was sick} 
\end{center}

 
\section{Morphological Segmentation Datasets}
\label{sec:msd}
To create our datasets, we make use of both segmentable (i.e., consisting of multiple morphemes) and
non-segmentable (i.e., consisting of one single morpheme) words described in books of the
collection \textit{Archive of Indigenous Languages} 
in Mexicanero \cite{canger2001mexicanero},
 Nahuatl \cite{lastra1980nahuatl}, 
 Wixarika \cite{gomez1999huichol},
 and Yorem Nokki \cite{freeze1989mayo}.
 Statistics about the data in the four languages are displayed in Tables \ref{tab:data3}, \ref{tab:data} and \ref{tab:data2}. 
We include segmentable as well as non-segmentable words into our datasets in order
to ensure that our methods can correctly decide against splitting up single morphemes. 
The phrases in all languages are mostly parallel, such that the corpora are roughly equivalent.
Therefore, we can compare the morphology of translated words (cf. Table \ref{tab:data2}), noticing that the language with most agglutination is Wixarika, with an average
rate of $3.25$ morphemes per word; the other languages have an average of close to $2.2$ morphemes per word. This higher morphological complexity 
naturally produces data sparsity at the token level. Also, we can notice that Wixarika has more unique words than the rest of our studied languages.
However, Nahuatl has with 810 the highest number of unique morphemes.  

\paragraph{Final splits.} 
\Katha{What are MorphTokens?}
\begin{table}[t]
    \setlength{\tabcolsep}{2.pt} 
  \centering
  \begin{tabular}{l || c | c | c | c }
            & Mex. & Nahuatl & Wixarika & Yorem N.\\\hline
      Words &       888  &    1123 & 1385     & 1063    \\
    SegWords&       539  &    746  & 1131     & 774     \\
      Morphs&      1889  &    2467 & 4502     & 2266    \\ 
  UniMorphs&       602  &    810  & 653      & 662     \\ \hline
      Seg/W &     0.606  &   0.664 & 0.816    & 0.728   \\
   Morphs/W &     2.127  &   2.196 & 3.250    & 2.131   \\
   MaxMorphs &         7  &       6 & 10       & 10      \\
  \end{tabular}
    \caption{Number of words, segmentable words (SegWords), total morphs (Morphs), and unique morphs (UniMorphs) in our datasets. Seg/W: proportion of words consisting or more than one morpheme; Morphs/W: morphemes per word; MaxMorphs: maximum number of morphemes found in one word.\label{tab:data2}}
\end{table} 
In order to make follow-up work on minimal-resource settings for morphological segmentation
easily comparable, we provide predefined splits of our
datasets\footnote{Our datasets can be found together with the code of our models at \texttt{http://turing.iimas.unam.mx/wix/MexSeg}.}. 
$40\%$ of the data constitute the test sets.
Of the remaining data, we use $20\%$ for development and the rest for training.
The final numbers of words per dataset and language are shown in Table \ref{tab:data}.

\section{Neural Seq2seq Models for Segmentation}
In the beginning of this section, we will introduce our neural architecture for segmentation. 
Subsequently, we will first describe our two proposed multi-task training approaches and second
our data augmentation methods. Finally, we will elaborate on expected differences between the two.

\subsection{Character-Based Encoder-Decoder RNN}
\label{subsec:ed_rnn}
Following work on segmentation by \newcite{kann-cotterell-schutze:2016:EMNLP2016} for high-resource settings,
our approach is based on the neural seq2seq model introduced by \newcite{bahdanau2015neural} for machine
translation.

\paragraph{Encoder.}
The first part of our model is a bidirectional recurrent neural network (RNN) 
which encodes the input sequence, i.e., the sequence of characters of a given word $w = w_1, w_2, \dots, w_{T_v}$, 
represented by the corresponding embedding vectors $v_{w_1}, ..., v_{w_{T_v}}$.
In particular, our encoder consists of one gated recurrent neural network (GRU) 
which processes the input in forward direction
and a second GRU which processes the input from the opposite side.

Encoding with this bidirectional GRU yields the forward hidden state
$\overrightarrow{h}_i = f\left(\overrightarrow{h}_{i-1},
v_i\right)$ and the backward hidden state $\overleftarrow{h}_i = f\left(\overleftarrow{h}_{i+1},
v_i\right)$, for a non-linear activation function $f$.
Their concatenation $h_i =
\left[{\overrightarrow{h_i}}; {\overleftarrow{h_i}}\right]$ is
passed on to the decoder.

\paragraph{Decoder.}
The second part of our network, the decoder,
is a single GRU, defining a probability distribution over
strings in $(\Sigma \cup {\tt S})^*$, for an alphabet $\Sigma$ and a separation symbol ${\tt S}$: 
\begin{equation*}
  p_{\text{ED}}(c \mid w) = \prod_{t=1}^{T_c} p(c_t \mid c_1, 
  \ldots,
  c_{t-1}, w).
\end{equation*}
where $p(c_t \mid c_1, \ldots, c_{t-1}, w)$ is computed using an attention 
mechanism and an output softmax layer over $\Sigma\cup {\tt S}$.  

A more detailed description of the general attention-based encoder-decoder architecture can be found in the original
paper by \newcite{bahdanau2015neural}.

\section{Improving Neural Models for Segmentation}

\subsection{Multi-Task Training}
\label{subsec:MTT}
In order to leverage unlabeled data or even random strings during training, we define an autoencoding auxiliary task, which 
consists of 
encoding the input and decoding an output which is identical to the original string.

Then, our multi-task training objective is to maximize the joint log-likelihood of 
this auxiliary task and our segmentation main task:
\begin{align}
  {\cal L}({\boldsymbol \theta})\!=& \!\!\!\sum_{(w, c) \in {\cal T}} \!\!\!\!\log \ptheta\left(c \mid e(w) \right) \label{eq:ll} \\ 
  & + \text{ } \!\!\!\sum_{a \in {\cal A}} \text{ }\quad\!\!\!\! \log \ptheta (a \mid e(a)) \nonumber
\end{align}
$\cal T$ denotes the segmentation training data with examples consisting of a word $w$ and 
its segmentation $c$. $\cal A$ denotes either a
set of words in the language of the system or a set of random strings.
The function $e$ describes the encoder and depends on the model parameters $\boldsymbol\theta$,
which are shared across the two tasks.
For training, we use data from both sets at the same time and mark each example with an
additional, task-specific input symbol.

We treat the size of $\cal A$ as a hyperparameter which we optimize on the development set separately for each language.
Values we experiment with are $m$ times the amount of instances in the original training set, with $m \in \{1, 2, 4, 8\}$.\footnote{An exception is Yorem Nokki, for which we do not have enough unlabeled data available, such that we experiment only with $m \in \{1, 2\}$.}

There are multiple reasons why we expect multi-task training to improve the performance of the final model. First, multi-task training should act as a regularizer. Second, for our models, the segmentation task consists in large parts of learning to copy the input character sequence to the output. This, however, can be learned from any string and does not require annotated segmentation boundaries. Third, in the case of unlabeled data (i.e., not for random strings), we expect the character language model in the decoder to improve, since it is trained on additional data.

We denote models trained with multi-task training using unlabeled corpus data as \textbf{MTT-U} and models
trained with multi-task training using random strings as \textbf{MTT-R}.

\subsection{Data Augmentation}
\label{subsec:DA}
A second option to make use of unlabeled data or random strings is to extend the available training data with new examples
made from those. 
The main question to answer here is how to include the new data into the existing datasets.
We do this by building new training examples in a fashion similar to the multi-task setup. All newly created instances
are of the form 
\begin{align}
w \mapsto w
\end{align}
where either $w \in V$ with $V$ being the observed vocabulary of the language, e.g., words in a given unlabeled corpus,
or $w \in R$ with $R$ being a set of sequences of random characters from the alphabet $\Sigma$ of the language. 

Again, we treat the amount of additional training examples as a hyperparameter which we optimize on the development set separately for each language.
We explore $m$ times the amount of instances in the original training set, with $m \in \{1, 2, 4, 8\}$.\ivan{Could it be possible to show valid examples of the new examples per language?}

The reasons why we expect our data augmentation methods to lead to better segmentation models are similar to those for multi-task training.

We call models trained on datasets augmented with unlabeled corpus data or random strings \textbf{DA-U} 
or \textbf{DA-R}, respectively.

\subsection{Differences Between Multi-task Training and Data Augmentation}
\label{subsec:differences}
The difference between MTT-U (resp. MTT-R) and DA-U (resp. MTT-U) is a single
element in the input sequence (the one representing the task). However, this information enables
the model to handle each given instance correctly at inference time. 
As a result, it gets more robust against noisy data, which seems crucial for our way of using unlabeled corpora.
Consider, for example, the Nahuatl word $onemokokowaya$. Training on
\begin{align*}
onemokokowaya \mapsto onemokokowaya
\end{align*}
will make the model learn \emph{not} to segment words which consist of the morphemes $o, ne, mo, kokowa, ya$,
which should ultimately hurt performance. 
The multi-task approach, in contrast, mitigates this problem.

As a conclusion, we expect the data augmentation approach with \emph{unlabeled data} to not obtain outstanding performance,
but rather consider it an important and informative baseline for the corresponding multi-task approach.
Using \emph{random strings}, the difference between the multi-task and the data augmentation approaches is less obvious:
Real morphemes should appear rarely enough in the created random character sequences to avoid the negative effect which we expect
for corpus words. We thus assume that the performances of MTT-R and DA-R should be similar.

\section{Experiments}

\subsection{Data}
We apply our models to the datasets described in \S\ref{sec:msd}.
For the multi-task training and data augmentation using unlabeled data, we use (unsegmented) words 
from a 
 parallel corpus collected by \newcite{gutierrez2016axolotl} for Nahuatl and the closely related Mexicanero.
 For Wixarika we use data from \newcite{magerwixnlp} and for Yorem Nokki we use text 
 from \newcite{sep2Yaqui}.

\subsection{Baselines}
Now, we will describe the baselines we use to evaluate the overall performance of our approaches.

\paragraph{Supervised seq2seq RNN (S2S).}
As a first baseline, we employ a fully supervised neural model without data augmentation or multi-task training, i.e., an attention-based
encoder-decoder RNN \cite{bahdanau2015neural} which has been trained only on the available annotated data.

\paragraph{Semi-supervised MORFESSOR (MORF).}
We further compare to the semi-supervised version of MORFESSOR \cite{kohonen2010semi}, a well-known morphological segmentation system.
During training, we tune the hyperparameters for each language on the respective development set. The best performing model is applied to the test set. 

\paragraph{FlatCat (FC).} Our next baseline is FlatCat \cite{gronroos-EtAl:2014:Coling}, a variant of MORFESSOR. It consists of a hidden Markov model for segmentation. The states of the model correspond either to a word boundary and one of the four morph categories stem,
prefix, suffix, and non-morpheme. It can work in an unsupervised way, but, similar to the previous baseline, can make effective use of small amounts of labeled data.

\paragraph{CRF.} We further compare to a 
conditional random fields (CRF) \cite{lafferty2001conditional} model, in particular a strong discriminative model for segmentation by \newcite{ruokolainen2014}. 
It reduces the task to a classification problem with four classes: beginning of a morph, middle of a morph, end of a morph and single character morph.  
Training is again semi-supervised and the model was previously reported to obtain good results for small amounts of unlabeled data \cite{ruokolainen2014}, which makes it very suitable for our minimal-resource setting.

\subsection{Hyperparameters}
\label{subsec:hyperparameters}
\paragraph{Neural network parameters. }
All GRUs in both the encoder and the decoder have
100-dimensional hidden states. All embeddings are 300-dimensional.

For training, we use ADADELTA (Zeiler, 2012) with a minibatch size of
20. We initialize all weights to the identity matrix and biases to
zero (Le et al., 2015). All models are trained for a maximum of 200
epochs, but we evaluate after every 5 epochs and apply the best performing model at test time.
Our final reported results are averaged accuracies over 5 single training runs.

\paragraph{Optimizing the amount of auxiliary task data. }
\begin{figure*}[h!]
\begin{minipage}{0.49\textwidth}
\centering
\begin{adjustbox}{width=0.87\textwidth}
\begin{tikzpicture}
\begin{axis}[legend pos=south west,
axis lines=middle,
ymin=0,
x label style={at={(current axis.right of origin)},anchor=north, below=10mm},
title={\textit{\textbf{MTT - corpus data}}},
    xlabel=times labeled data,
  ylabel=\% accuracy,
  xticklabel style = {rotate=30,anchor=east},
   enlargelimits = false,
  xticklabels from table={MTL-C.dat}{Time},xtick=data,
  ymin=40,ymax=90]
\addplot[orange,thick,mark=square*] table [y=Mexicanero,x=X]{MTL-C.dat};
\addlegendentry{Mexicanero}
\addplot[yellow,thick,mark=square*] table [y={Yorem Nokki},x=X]{MTL-C.dat};
\addlegendentry{Yorem Nokki}
\addplot[green,thick,mark=square*] table [y=Nahuatl,x=X]{MTL-C.dat};
\addlegendentry{Nahuatl}
\addplot[blue,thick,mark=square*] table [y=Wixarika,x=X]{MTL-C.dat};
\addlegendentry{Wixarika}]
\end{axis}
\end{tikzpicture}
\end{adjustbox}
\end{minipage}
\hfill
\begin{minipage}{0.49\textwidth}
\centering
\begin{adjustbox}{width=0.87\textwidth}
\begin{tikzpicture}
\begin{axis}[legend pos=south west,
axis lines=middle,
ymin=0,
x label style={at={(current axis.right of origin)},anchor=north, below=10mm},
title={\textit{\textbf{MTT - random strings}}},
    xlabel=times labeled data,
  ylabel=\% accuracy,
  xticklabel style = {rotate=30,anchor=east},
   enlargelimits = false,
  xticklabels from table={MTL-R.dat}{Time},xtick=data,
  ymin=40,ymax=90]
\addplot[orange,thick,mark=square*] table [y=Mexicanero,x=X]{MTL-R.dat};
\addlegendentry{Mexicanero}
\addplot[yellow,thick,mark=square*] table [y={Yorem Nokki},x=X]{MTL-R.dat};
\addlegendentry{Yorem Nokki}
\addplot[green,thick,mark=square*] table [y=Nahuatl,x=X]{MTL-R.dat};
\addlegendentry{Nahuatl}
\addplot[blue,thick,mark=square*] table [y=Wixarika,x=X]{MTL-R.dat};
\addlegendentry{Wixarika}]
\end{axis}
\end{tikzpicture}
\end{adjustbox}
\end{minipage}
\caption{Accuracy on the development set in dependence of the amount of auxiliary task training data for multi-task learning.}
\label{Fig:MTT}
\end{figure*}
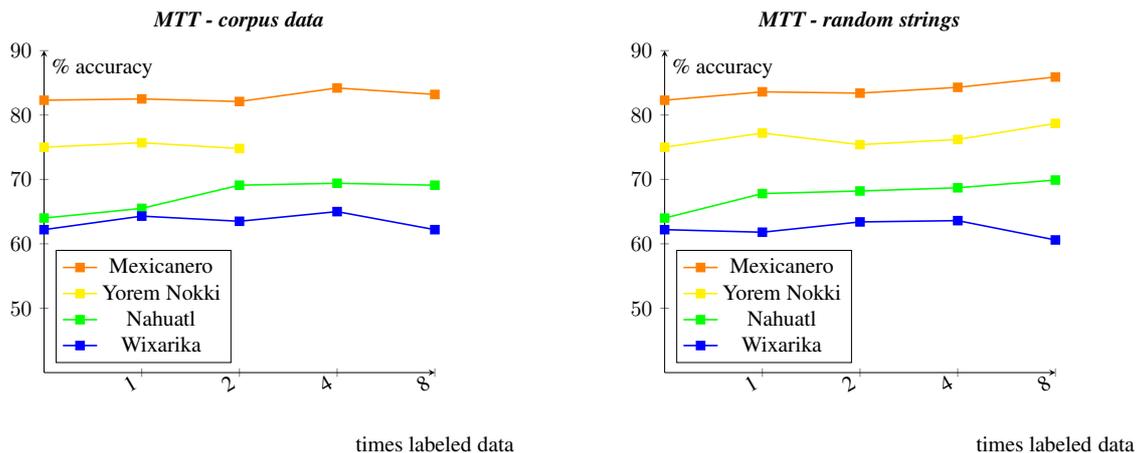

\begin{figure*}[h!]
\begin{minipage}{0.49\textwidth}
\centering
\begin{adjustbox}{width=0.87\textwidth}
\begin{tikzpicture}
\begin{axis}[legend pos=south west,
axis lines=middle,
ymin=0,
x label style={at={(current axis.right of origin)},anchor=north, below=10mm},
title={\textit{\textbf{DA - corpus data}}},
    xlabel=times labeled data,
  ylabel=\% accuracy,
  xticklabel style = {rotate=30,anchor=east},
   enlargelimits = false,
  xticklabels from table={DA-C.dat}{Time},xtick=data,
  ymin=40,ymax=90]
\addplot[orange,thick,mark=square*] table [y=Mexicanero,x=X]{DA-C.dat};
\addlegendentry{Mexicanero}
\addplot[yellow,thick,mark=square*] table [y={Yorem Nokki},x=X]{DA-C.dat};
\addlegendentry{Yorem Nokki}
\addplot[green,thick,mark=square*] table [y=Nahuatl,x=X]{DA-C.dat};
\addlegendentry{Nahuatl}
\addplot[blue,thick,mark=square*] table [y=Wixarika,x=X]{DA-C.dat};
\addlegendentry{Wixarika}]
\end{axis}
\end{tikzpicture}
\end{adjustbox}
\end{minipage}
\hfill
\begin{minipage}{0.49\textwidth}
\centering
\begin{adjustbox}{width=0.87\textwidth}
\begin{tikzpicture}
\begin{axis}[legend pos=south west,
axis lines=middle,
ymin=0,
x label style={at={(current axis.right of origin)},anchor=north, below=10mm},
title={\textit{\textbf{DA - random strings}}},
    xlabel=times labeled data,
  ylabel=\% accuracy,
  xticklabel style = {rotate=30,anchor=east},
   enlargelimits = false,
  xticklabels from table={DA-R.dat}{Time},xtick=data,
  ymin=40,ymax=90]
\addplot[orange,thick,mark=square*] table [y=Mexicanero,x=X]{DA-R.dat};
\addlegendentry{Mexicanero}
\addplot[yellow,thick,mark=square*] table [y={Yorem Nokki},x=X]{DA-R.dat};
\addlegendentry{Yorem Nokki}
\addplot[green,thick,mark=square*] table [y=Nahuatl,x=X]{DA-R.dat};
\addlegendentry{Nahuatl}
\addplot[blue,thick,mark=square*] table [y=Wixarika,x=X]{DA-R.dat};
\addlegendentry{Wixarika}]
\end{axis}
\end{tikzpicture}
\end{adjustbox}
\end{minipage}
\caption{Accuracy on the development set in dependence of the amount of additional training data.}
\label{Fig:DA}
\end{figure*}
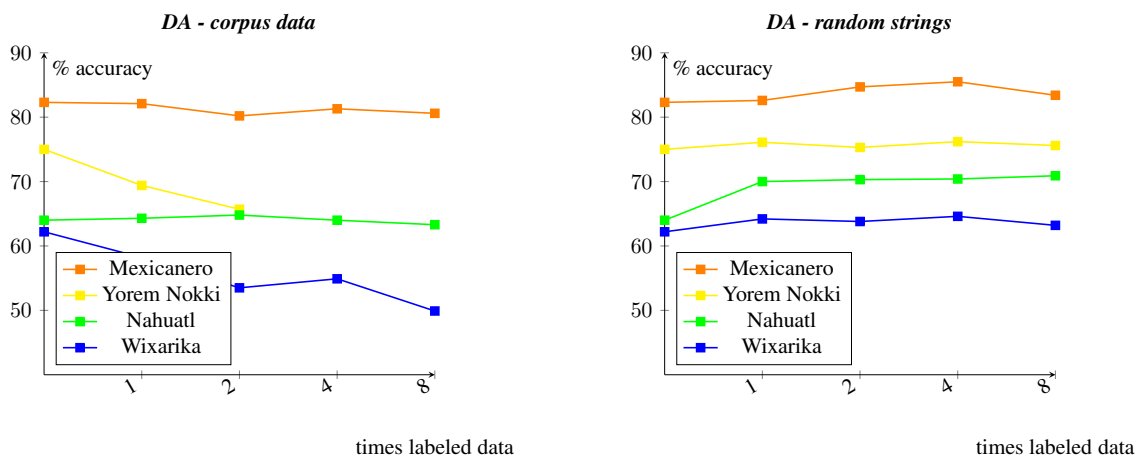
The performance of our neural segmentation model in dependence of the amount of auxiliary task training data can be 
seen in Figure \ref{Fig:MTT}.
As a general tendency across all languages, adding more data seems better, particularly for the 
autoencoding task with random strings. The only exception is Wixarika. 

The final configurations we choose for $m$ (cf. \S\ref{subsec:MTT}) in the case of multi-task training with 
the auxiliary task of autoencoding corpus data are $m=4$ for 
Mexicanero, Nahuatl and Wixarika and $m=1$ for Yorem Nokki. For multi-task training with autoencoding of random strings
we select $m=8$ for Mexicanero, Nahuatl and Yorem Nokki and $m=4$ for Wixarika.

\paragraph{Optimizing the amount of artificial training data for data augmentation. }
Figure \ref{Fig:DA} shows the performance of the encoder-decoder depending on the amount of added artificial training data.
In the case of random strings, again, adding more training data seems to help more. However, using corpus data 
seems to hurt performance and the more such examples we use, the worse accuracy we obtain. Thus, we conclude that 
(as expected) data augmentation with corpus 
data is not a good way to improve the model's performance. We will discuss this in more detail in \S\ref{subsec:results}.

Even though the final conclusion should be to not add much corpus data,
we apply what gives best results on the development set.
The final configurations we thus choose for DA-U are $m=1$ for 
Mexicanero, Wixarika and Yorem Nokki and $m=2$ for Nahuatl. 
For DA-R, we select $m=4$ for 
Mexicanero, Wixarika and Yorem Nokki and $m=8$ for Nahuatl.

\subsection{Evaluation Metrics}

\paragraph{Accuracy.} First, we evaluate using accuracy on the token level. Thus, an example counts as correct if and only if the output of the system matches the reference solution exactly, i.e., if all output symbols are predicted correctly.

\paragraph{F1.} Our second evaluation metric is border
F1, which measures how many segment boundaries are predicted correctly by the model.
While we use this metric because it is common for segmentation tasks, it is not ideal for our models since those are not guaranteed to preserve the input character sequence.
We handle this problem as follows: In order to compare borders, we identify them by the position of their preceding letter, i.e., 
if in both the model's guess and the gold solution a segment border appears after the second character, it counts as correct. Wrong characters are ignored. Note that this comes with the disadvantage of erroneously inserted characters leading to all subsequent segment borders being counted as incorrect.

\subsection{Test Results and Discussion}
\label{subsec:results}
\begin{table*}[]
    \setlength{\tabcolsep}{1.2pt} 
  \small
  \centering
  \begin{tabular}{l || c | c | c | c | c | c | c | c  || c | c | c | c | c | c | c | c} 
    & \multicolumn{8}{c ||}{Accuracy} & \multicolumn{8}{c}{F1} \\
    & MTT-U & MTT-R & DA-U & DA-R & S2S & MORF & CRF & FC & MTT-U & MTT-R & DA-U & DA-R & S2S & MORF & CRF & FC \\\hline 
    Mex. & \textbf{.8051} &  .7955 & .7611 & .7983 & .7504 & .3364 & .7837 & .5420
    & \textbf{.8786} & .8694 & .6715 & .8683 & .8618 & .5121 & .8639 & .5621 \\ 
    Nahuatl & .6004 & .6027 & .5541 & .6018 &  .5585 & .4044 & \textbf{.6444} & .4888
    & .7388 & .7367 & .6865 & .7328 & .7266 & .4154 & \textbf{.7487} & .5185 \\
    Wixarika & .5895 & .6134 & .5425 & \textbf{.6188} &  .5754 & .3989 & .5866 & .4523
    & .7949 & .8024 & .7109 & \textbf{.8161} & .7961  & .4426 & .7932 & .5568 \\
    Yorem N. & .6856 & \textbf{.7101} & .6212 & .6936 & .6569 & .4812 & .6596 & .5781
    & .7887 & \textbf{.8076} & .7133 & .7923 & .7730 & .3528 & .7736 & .6139 \\
  \end{tabular}
  \caption{Performances of our multi-task and data augmentation approaches compared to all baselines described in the text. The reported results for neural models are averages over 5 training runs. Best results per language and metric are in bold.\label{results}}
\end{table*} 
Table \ref{results} shows that accuracy and F1 seem to be highly correlated for our task. The test results also give an answer to our first research question: The neural model S2S performs on par with CRF,
the strongest baseline, for all languages but Nahuatl.
Further, S2S and CRF both outperform MORF and FC by a wide margin. We may thus conclude that neural models are indeed applicable to segmentation of polysynthetic languages in a low-resource setting.

Second, we can see that all our proposed methods except for DA-U improve over
S2S, the neural baseline:
The accuracy of MTT-U is between $0.0141$ (Wixarika) and $0.0547$ (Mexicanero) higher than S2S's.
MTT-R improves between $0.0380$ (Wixarika) and
$0.0532$ (Yorem Nokki). Finally, DA-R outperforms S2S by 
$0.0367$ to $0.0479$ accuracy for Yorem Nokki and Mexicanero, respectively.
The overall picture when considering F1 looks similar.
Comparing our approaches to each other, there is no clear winner. This might be due to differences in the unlabeled 
data we use: the corpus we use for Mexicanero and Nahuatl is from dialects different from both respective test sets. 
Assuming that the effect of training a language model using unlabeled data and erroneously learning to not segment words 
are working against each other for MTT-U, this might explain why MTT-U is best for Mexicanero and the gap between MTT-U and MTT-R is
smaller for Nahuatl than for Yorem Nokki and Wixarika.

As mentioned before (cf. \S\ref{subsec:differences}), a simple data augmentation method using unlabeled data should 
hurt performance. This is indeed the result of our experiments: DA-U performs worse than S2S for all languages except for Mexicanero,
where the unlabeled corpus is from another language: the closely related Nahuatl.
We thus conclude that multi-task training (instead of simple data augmentation) is crucial for the use of 
unlabeled data.

Finally, our methods compare favorably to all baselines, with the exception of CRF for Nahuatl. While CRF is overall the strongest baseline for 
our considered languages, our methods outperform it by up to $0.0214$ accuracy or $0.0147$ F1 for Mexicanero, $0.0322$ accuracy or $0.0229$ F1 for Wixarika and $0.0505$ accuracy or $0.0340$ F1 for Yorem Nokki.
This shows the effectiveness of our fortified neural models for minimal-resource morphological segmentation.

\section{Cross-Lingual Transfer Learning}
We now want to investigate the performance of \emph{one single model} 
trained on all languages at once. 
This is done in analogy to the multi-task training described in \S\ref{subsec:MTT}.
We treat segmentation in each language as a separate task and train an attention-based encoder-decoder model on maximizing
the joint log-likelihood:
\begin{align}
  {\cal L}({\boldsymbol \theta})\!=& \!\!\!\sum_{L_i \in L} \sum_{(w, c) \in {{\cal T}_{L_i}}} \!\!\!\!\log \ptheta\left(c \mid e(w) \right) \nonumber\\ 
\end{align}
${\cal T}_{L_i}$ denotes the segmentation training data in language $L_i$ and $L$ is the set of our languages. As before, each training example consists
of a word $w$ and its segmentation $c$.

\subsection{Experimental Setup}
We keep all model parameters and the training regime as described in \S\ref{subsec:hyperparameters}.
However, our training data now consists of a combination of all available training data for all 4 languages.
In order to enable the model to differentiate between the tasks, we prepend one language-specific input symbol to each instance.
This corresponds to having one embedding in the input which marks the task. 
An example training instance for Yorem Nokki is
\begin{align*}
\text{\textit{L=YN}  }ko'koreyene \mapsto ko'kore|ye|ne,
\end{align*}
where \textit{L=YN} indicates the language.

Due to the previous high correlation between accuracy and F1 
we only use accuracy on the word level as the evaluation metric for this experiment.

\subsection{Results and Discussion}
\begin{table}[]
    \setlength{\tabcolsep}{2.5pt} 
  \centering
  \begin{tabular}{l || c | c | c | c } 
    & M-Lang & S-Lang & BestMTT & BestDA \\\hline 
    Mex. & .6858 & .7504 & \textbf{.8051} & .7983  \\ 
    Nahuatl & .5955 & .5585 & \textbf{.6027} & .6018  \\
    Wixarika & .6021 & .5754 & .6134 & \textbf{.6188} \\
    Yorem N. & .6223 & .6569 & \textbf{.7101} & .6936 \\ 
  \end{tabular}
  \caption{Accuracies of our model trained on all languages (M-Lang) and the models trained on single languages (S-Lang). The highest multi-task and data augmentation 
  accuracies are repeated for an easy comparison.\label{results_crossling}}
\end{table} 
In Table \ref{results_crossling}, we show the results of the multi-lingual model, which was trained on all languages, compared to all individual models, as well as each respective best multi-task approach and data augmentation method. 
The results differ among languages: 
Most remarkably, for both Wixarika and Nahuatl, the accuracy of the multi-lingual model is higher than the one of the single-language model. 
This might be related to them being the languages with most training data available (cf. Table \ref{tab:data2}).

Note, however, that even for the remaining two languages---Mexicanero and Yorem Nokki---we hardly lose accuracy when comparing the multi-lingual to the individual models. 
Since we only use one model (instead of four), without increasing its size significantly, we thus reduce the amount of parameters by nearly $75\%$.

\section{Related Work}
Work on morphological segmentation was started more than 6 decades ago \cite{harris1951methods}.
Since then, many approaches have been developed: 
In the realm of unsupervised methods, two important systems are LINGUISTICS \cite{Goldsmith:2001:ULM:972667.972668} and
MORFESSOR \cite{creutz-lagus:2002:ACL02-MPL}. The latter was later extended to a semi-supervised version \cite{kohonen2010semi} in order to
make use of the abundance of unlabeled data which is available for many languages.

\newcite{ruokolainen-EtAl:2013:CoNLL-2013} focused explicitly on low-resource scenarios and applied 
CRFs to morphological segmentation in several languages. They reported better results than earlier work, including semi-supervised approaches. In the following year, they extended their approach to be able to use 
unlabeled data as well, further improving performance \cite{ruokolainen2014}. 

\newcite{cotterell-EtAl:2015:CoNLL} trained a semi-Markov CRF (semi-CRF) \cite{sarawagi2005semi} jointly on morphological segmentation, stemming and tagging. For the similar problem of Chinese word 
segmentation, 
\newcite{zhang2008joint} trained a model jointly on part-of-speech tagging. 
However, we are not aware of any prior work on 
multi-task training or data augmentation for neural segmentation models.

In fact, the two only neural seq2seq approaches for morphological segmentation we know of 
focused on \emph{canonical segmentation} \cite{cotterell-vieira-schutze:2016:N16-1} which differs from
the \emph{surface segmentation} task considered here in that it restores changes to the surface form of morphemes which occurred during word formation. 
\newcite{kann-cotterell-schutze:2016:EMNLP2016} also used an encoder-decoder RNN and combined it with a neural reranker. While our 
model architecture was inspired by them, their model was purely supervised. Additionally, they did not investigate the applicability of their neural seq2seq model in low-resource settings or for polysynthetic languages. \newcite{ruzsics-samardzic:2017:CoNLL} extended the standard encoder-decoder architecture for canonical segmentation to contain a language model over segments and improved results. However,
a big difference to our work is that they still used more than ten times as much training data as we have available for the indigenous Mexican languages we are working on here.

Another neural approach---this time for surface segmentation---was presented by \newcite{wang2016morphological}. The authors, instead of using seq2seq models, treat the task as a sequence labeling problem and use LSTMs to classify every character either as the beginning, middle or end of a morpheme, or as a single-character morpheme.

Cross-lingual knowledge transfer via language tags was proposed for neural seq2seq models before, both for tasks that handle sequences of words \cite{TACL1081} and tasks that work on sequences of characters \cite{kann-cotterell-schutze:2017:Long}.
However, to the best of our knowledge, we are the first to try such an approach for a morphological segmentation task. 
In many other areas of NLP, cross-lingual transfer has been applied successfully, e.g., in entity
recognition \cite{MengqiuWang2014}, language modeling \cite{tsvetkov-EtAl:2016:N16-1}, or parsing
\cite{CohenDS11,sogaard:2011:ACL-HLT20112,TACL892}.

\section{Conclusion and Future Work}
We first investigated the applicability of neural seq2seq models 
to morphological surface segmentation for polysynthetic languages in minimal-resource settings,
i.e., for considerably less than $1,000$ training instances.
Although they are generally thought to require
large amounts of training data, neural networks obtained an accuracy comparable
to or higher than several strong baselines.

Subsequently, we proposed two novel multi-task training approaches and two novel data augmentation methods to further increase  
the performance of our neural models. 
Adding those, we improved over the neural baseline for all languages, and for Mexicanero, Wixarika and Yorem Nokki our final models outperformed all baselines by up to $5.05\%$ absolute accuracy or $3.40\%$ F1.
Furthermore, we explored  cross-lingual transfer between our languages and reduced the amount of necessary model parameters by about $75\%$, while improving performance for some of the languages.

We publically 
release our datasets for morphological surface
segmentation of the polysynthetic minimal-resource languages Mexicanero,
Nahuatl, Wixarika and Norem Yokki.

\section*{Acknowledgments}
We would like to thank Paulina Grnarova, Rodrigo Nogueira and Ximena Gutierrez-Vasques for their helpful feedback.

\bibliography{naaclhlt2018}
\bibliographystyle{acl_natbib}

\end{document}